\documentclass{article}

\usepackage{microtype}
\usepackage{graphicx}
\usepackage{booktabs}
\usepackage{multirow}
\usepackage{amsmath}
\usepackage{amssymb}
\usepackage{hyperref}
\usepackage[table]{xcolor}
\usepackage{colortbl}

\definecolor{darkgreen}{RGB}{0,120,0}
\definecolor{darkred}{RGB}{160,0,0}

\usepackage[accepted]{icml2026}

\icmltitlerunning{Interpretable Sperm Morphology Classification via Attention-Guided Deep Learning}

\begin{document}

\twocolumn[
  \icmltitle{Interpretable Sperm Morphology Classification\\
             via Attention-Guided Deep Learning}

  \icmlsetsymbol{equal}{*}

  \begin{icmlauthorlist}
    \icmlauthor{Zahra Asghari Varzaneh}{mau}
    \icmlauthor{Reza Khoshkangini}{mau}
    \icmlauthor{Thomas Ebner}{kuk}\icmlauthor{ Lars Johansson}{AB} 
    
  \end{icmlauthorlist}

  \icmlaffiliation{mau}{Department of Computer Science and Media Technology,
    Malmö University, Malmö, Sweden}
  \icmlaffiliation{kuk}{Kepler Universitätsklinikum, Linz, Austria}
 \icmlaffiliation{AB}
  {NewLifeAid-Global AB, Sweden}

  \icmlcorrespondingauthor{Zahra Asghari Varzaneh}{zahra.asghari-varzaneh@mau.se}

  \icmlkeywords{sperm morphology, deep learning, EfficientNet, CBAM attention, Grad-CAM++, explainability}

  \vskip 0.3in
]

\printAffiliationsAndNotice{}

%-------------------------------------------------------------------
\begin{abstract}
Male infertility is a major cause of couple infertility, often linked to abnormal sperm morphology. While deep learning models offer automated analysis, most lack interpretability, limiting their clinical adoption. This study proposes an attention-guided deep learning framework for sperm morphology classification. We combine a pretrained EfficientNet-B0 with a Convolutional Block Attention Module (CBAM) to focus on key areas of the sperm head, improving both accuracy and interpretability. Evaluated on the SMIDS and HuSHem public datasets, our model achieves accuracies of 90.2\% and 93.9\% (macro F1-scores of 0.913 and 0.948), outperforming SimpleCNN and standard EfficientNet-B0. Furthermore, we use Grad-CAM++ visualizations to highlight features influencing the model's decisions. The results demonstrate that this accurate and transparent framework is a practical tool for automated sperm analysis in fertility clinics.
\end{abstract}

%-------------------------------------------------------------------
\section{Introduction}
\label{sec:intro}
Infertility is a global health issue affecting nearly 15\% of couples, with male factors contributing to about half of these cases. Around 30\% of male infertility cases relate directly to poor sperm quality and abnormal parameters \cite{Agarwal2015}. Various biological, environmental, and lifestyle factors can negatively impact sperm count, motility, and morphology, reducing pregnancy chances \cite{Bonde1993,Durairajanayagam2018}. Consequently, semen analysis is the primary test for evaluating male fertility, where sperm morphology assessment plays a critical role in predicting reproductive outcomes \cite{Patel2018}. Usually, this process is performed manually by laboratory specialists using light microscopy based on standard clinical guidelines \cite{Kruger1988}.

However, manual analysis is time-consuming, subjective, and highly dependent on the observer's experience. Studies show that observer agreement can be as low as 60--70\% \cite{Tomlinson2010}. This lack of consistency, especially during high daily workloads, increases human error and limits diagnostic reliability. Therefore, objective and reproducible automated methods for sperm morphology analysis are highly needed.\\
Recent advances in artificial intelligence and deep learning (DL) have shown strong performance in medical image processing \cite{Varzaneh2025a,Zamani2025,Varzaneh2025b}. Recently, these techniques have been applied to sperm morphology classification. For instance, Liu et al. \cite{Liu2021} used a modified AlexNet with transfer learning to classify human sperm morphology and reduce computational costs. Similarly, Cansiz et al. \cite{Cansiz2025} introduced a generative adversarial framework to improve classification accuracy. Despite such progress, most DL models operate as "black boxes," providing predictions without explaining the underlying decision process. This lack of transparency limits their adoption in clinical practice \cite{Sendak2020}.
To resolve this issue, explainable artificial intelligence (XAI) methods like Grad-CAM++ \cite{Chattopadhay2018} have been developed. These techniques generate visual heatmaps to highlight key regions, helping clinicians verify model decisions.\\
In this study, we propose an attention-guided deep learning framework for sperm morphology classification. Our model combines EfficientNet-B0 \cite{Tan2019} with the Convolutional Block Attention Module (CBAM) \cite{Woo2018} to focus on the most important regions of the sperm head. We evaluate our method on the SMIDS and HuSHem datasets, comparing it against a custom CNN and standard EfficientNet-B0. Additionally, Grad-CAM++ is used for visual explanations. The main contributions of this work are:\\
• Proposing an EfficientNet-B0 + CBAM framework for accurate and interpretable sperm morphology classification.\\
• Integrating a freeze-then-unfreeze training strategy to enhance performance on small datasets like HuSHem.\\
• Generating class-specific visual explanations using Grad-CAM++ to improve clinical trust.

%-------------------------------------------------------------------
\begin{figure*}[t]
\centering
\includegraphics[width=0.95\textwidth]{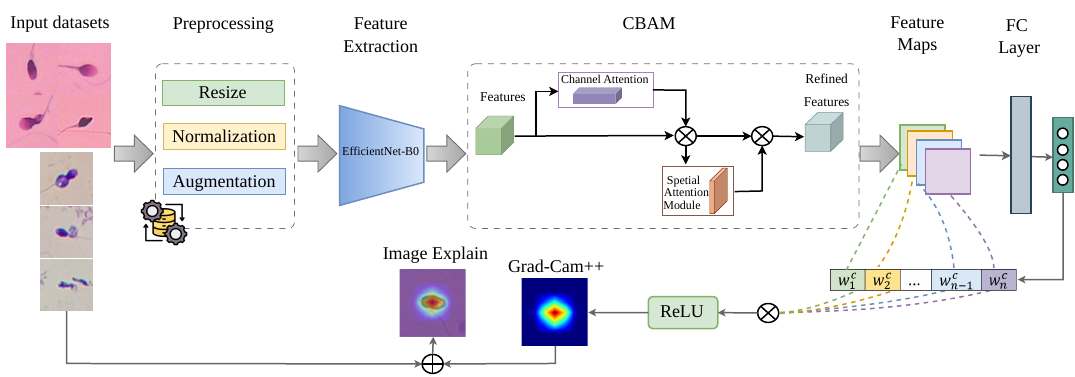}
\caption{Overall architecture of the proposed framework.}
\label{fig:framework}
\end{figure*}

\section{Methodology}
\label{sec:method}
We propose an attention-guided deep learning framework for automatic sperm morphology classification, combining EfficientNet-B0 for feature extraction with CBAM to focus on diagnostically important regions of the sperm head. To ensure clinical transparency, Grad-CAM++ is used for post-hoc visualization. As shown in Figure 1, the overall pipeline consists of four main stages: (1) data preprocessing and augmentation, (2) feature extraction via EfficientNet-B0 + CBAM, (3) training using a freeze-then-unfreeze strategy for small datasets, and (4) visual explanation via Grad-CAM++.

\subsection{Preprocessing and Augmentation}
Images are resized to $224 \times 224$ pixels and normalized using ImageNet statistics. Training augmentations include random flipping, rotation, color jittering, and affine transformations. For the small HuSHem dataset, MixUp augmentation (Eq.~\ref{eq:1}) and label smoothing are applied as regularization strategies:
\begin{equation}
\tilde{x} = \lambda x_i + (1-\lambda)x_j
\label{eq:1}
\end{equation}
where $\lambda$ is sampled from a Beta distribution.

\subsection{EfficientNet-B0 Backbone}
We use a compact pretrained EfficientNet-B0~\cite{Tan2019} initialized with ImageNet weights to leverage its compound scaling method. Given an input image $X$, the backbone extracts deep feature representations $F \in \mathbb{R}^{C \times H \times W}$ via:
\begin{equation}
F = f_{\theta}(X)
\label{eq:2}
\end{equation}
The resulting 1,280-channel feature map is then passed to the CBAM module.

\subsection{CBAM Attention Module}
To prioritize key features, CBAM~\cite{Woo2018} sequentially applies channel and spatial attention. The channel-attended feature map $F'$ is computed as:
\begin{equation}
F' = M_c(F) \otimes F
\label{eq:3}
\end{equation}
Next, the spatial attention module refines these features focusing on critical regions:
\begin{equation}
F'' = M_s(F') \otimes F'
\label{eq:4}
\end{equation}
where $M_c$ and $M_s$ represent channel and spatial attention maps, and $\otimes$ is element-wise multiplication.

\subsection{Freeze-then-Unfreeze Training}
To prevent overfitting on small data, a two-stage strategy is used. First, the backbone and CBAM layers are frozen while only the classification head is trained. Second, all parameters are fine-tuned jointly with a lower learning rate using cross-entropy loss:
\begin{equation}
L_{CE} = -\sum_{i=1}^{K} y_i \log(\hat{y}_i)
\label{eq:5}
\end{equation}
where $K$ is the class count, $y_i$ is the true label, and $\hat{y}_i$ is the predicted probability. Optimization is done via AdamW with cosine annealing.

\subsection{Grad-CAM++ for Explainability}
Grad-CAM++~\cite{Chattopadhay2018} generates visual heatmaps to highlight predictive regions. Class-specific importance weights are calculated from the final convolutional feature maps using:
\begin{equation}
L^{c}_{GradCAM++} = \mathrm{ReLU}\left(\sum_{k} w_k^c A^k \right)
\label{eq:6}
\end{equation}
where $A^k$ is the $k$-th feature map and $w_k^c$ is its weight for class $c$.

\section{Experimental Setup and Datasets}
We evaluate the framework on two public datasets. \textbf{SMIDS}~\cite{Ilhan2018} contains 3,000 microscopic images categorized into three classes: Normal, Abnormal, and non-sperm. \textbf{HuSHem}~\cite{Shaker2017} contains 216 expert-verified sperm head images across four classes: Normal, Tapered, Pyriform, and Amorphous. \\
Experiments are implemented in PyTorch 2.6 on an NVIDIA GPU with a batch size of 32 for 100 epochs. Data is split into 70\% training, 15\% validation, and 15\% testing using a fixed random seed. We compare three models: (1) a baseline SimpleCNN trained from scratch, (2) standard pretrained EfficientNet-B0, and (3) our proposed EfficientNet-B0 + CBAM model.

%-------------------------------------------------------------------
\section{Results and Discussion}

\begin{figure}[t]
\centering
\includegraphics[width=1.06\linewidth]{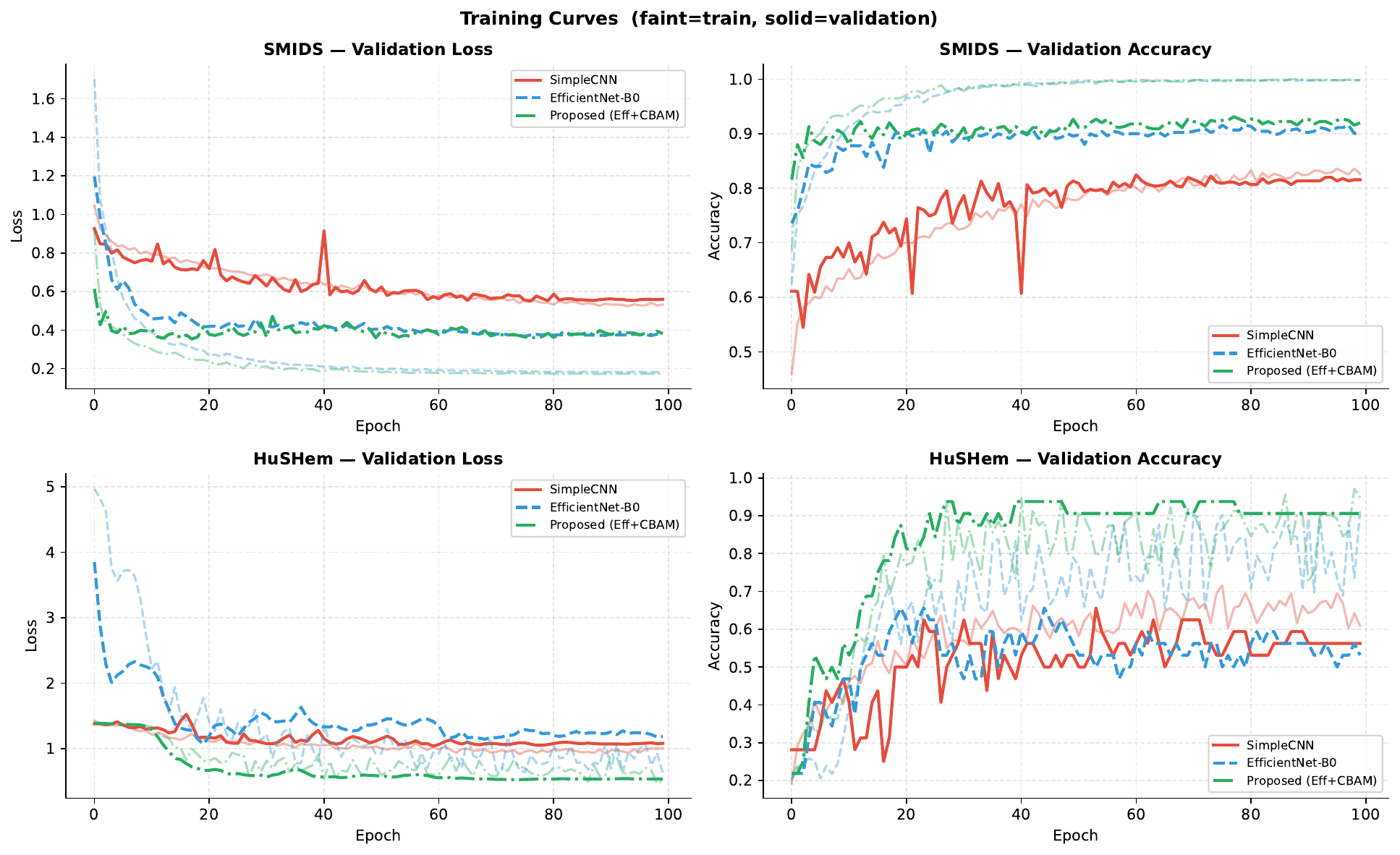}
\caption{Training and validation curves for SMIDS and HuSHem datasets. The proposed model shows stable convergence and strong generalization. The two-phase behavior on HuSHem reflects the freeze-then-unfreeze strategy.}
\label{fig:train}
\end{figure}

\begin{figure}[t]
\centering
\includegraphics[width=1.02\linewidth]{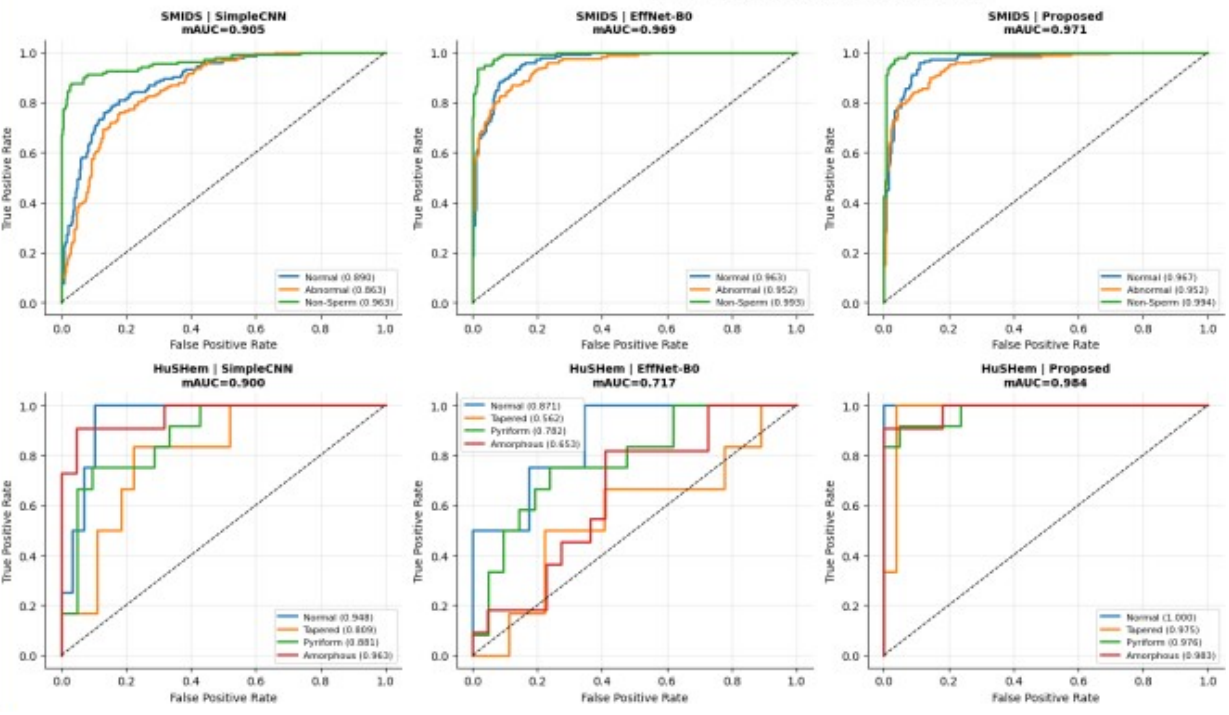}
\caption{ROC curves for all models on SMIDS and HuSHem datasets using a one-vs-rest evaluation scheme.}
\label{fig:roc}
\end{figure}

Table~\ref{tab:results} summarizes accuracy and macro F1-score on the test sets for both datasets. The proposed method consistently achieves the best performance. On SMIDS, it reaches 90.2\% accuracy and F1 = 0.913, outperforming both SimpleCNN and EfficientNet-B0 baseline while also providing interpretability through CBAM attention and Grad-CAM++ visualizations. On HuSHem, the proposed method achieves 93.9\% accuracy and F1 = 0.948, compared to 63.6\% for EfficientNet-B0 and 72.7\% for SimpleCNN.\\
This substantial performance gap on HuSHem demonstrates that the CBAM attention mechanism is especially effective for small datasets, as it forces the model to focus on morphologically relevant sperm head regions rather than background noise. The results also confirm the effectiveness of the freeze-then-unfreeze training strategy for medical image classification.
\begin{table}[b]
\centering
\footnotesize
\setlength{\tabcolsep}{3pt}
\caption{Classification results on SMIDS and HuSHem test sets. Best results in bold.}
\label{tab:results}
\begin{tabular}{lcccc}
\hline
Model & \multicolumn{2}{c}{SMIDS} & \multicolumn{2}{c}{HuSHem} \\
\cline{2-5}
 & Accuracy & F1-score & Accuracy & F1-score \\
\hline
SimpleCNN & 82.67\% & 0.830 & 72.73\% & 0.695 \\
EfficientNet-B0 & 88.00\% & 0.883 & 63.64\% & 0.684 \\
Proposed & \textbf{90.21\%} & \textbf{0.913} & \textbf{93.94\%} & \textbf{0.948} \\
\hline
\end{tabular}
\end{table}
To provide a more detailed analysis, Table~\ref{tab:perclass} reports the per-class performance. On SMIDS, the proposed model achieves the highest performance for the Non-Sperm class (F1=0.97), which is clinically important because misclassifying debris as sperm can affect sperm count estimation. The Normal and Abnormal classes show balanced performance, indicating no strong bias toward a specific class.\\
On HuSHem, the proposed framework achieves strong discrimination across all morphology categories, with near-perfect performance on the Normal class and consistently high scores on Tapered, Pyriform, and Amorphous classes. This confirms that the attention-guided framework effectively captures subtle morphological differences in sperm head shapes.
\begin{table}[b]
\centering
\scriptsize
\setlength{\tabcolsep}{4pt} % کمی فاصله ستون‌ها بیشتر شد تا خواناتر شود
\renewcommand{\arraystretch}{1.2} % افزایش جزئی فاصله عمودی سطرها برای زیبایی بیشتر

\caption{Per-class performance comparison (Precision \textbar{} Recall \textbar{} F1-score) for all models on SMIDS and HuSHem datasets.}
\label{tab:perclass}

\begin{tabular}{llccc}
\toprule
Dataset & Class & SimpleCNN & EfficientNet-B0 & Proposed \\
\midrule
\multirow{3}{*}{SMIDS}
& Normal    & 0.78 \textbar{} 0.85 \textbar{} 0.81 & 0.84 \textbar{} 0.88 \textbar{} 0.86 & 0.88 \textbar{} 0.88 \textbar{} 0.88 \\
& Abnormal  & 0.81 \textbar{} 0.74 \textbar{} 0.77 & 0.86 \textbar{} 0.82 \textbar{} 0.84 & 0.87 \textbar{} 0.87 \textbar{} 0.87 \\
& Non-Sperm & 0.91 \textbar{} 0.91 \textbar{} 0.91 & 0.95 \textbar{} 0.95 \textbar{} 0.95 & 0.97 \textbar{} 0.96 \textbar{} 0.97 \\
\midrule
\multirow{4}{*}{HuSHem}
& Normal    & 0.40 \textbar{} 1.00 \textbar{} 0.57 & 1.00 \textbar{} 1.00 \textbar{} 1.00 & 1.00 \textbar{} 1.00 \textbar{} 1.00 \\
& Tapered   & 0.75 \textbar{} 0.50 \textbar{} 0.60 & 0.50 \textbar{} 0.83 \textbar{} 0.62 & 0.86 \textbar{} 1.00 \textbar{} 0.92 \\
& Pyriform  & 1.00 \textbar{} 0.58 \textbar{} 0.74 & 0.67 \textbar{} 0.67 \textbar{} 0.67 & 0.92 \textbar{} 0.92 \textbar{} 0.92 \\
& Amorphous & 0.83 \textbar{} 0.91 \textbar{} 0.87 & 0.57 \textbar{} 0.36 \textbar{} 0.44 & 1.00 \textbar{} 0.91 \textbar{} 0.95 \\
\bottomrule
\end{tabular}
\end{table}
Figure~\ref{fig:train} illustrates the training and validation curves of the proposed model on both datasets. On SMIDS, the model converges rapidly and remains stable, indicating strong generalization. On HuSHem, the effect of the freeze-then-unfreeze strategy is clearly visible: validation accuracy increases gradually during the frozen stage and improves significantly after unfreezing the backbone, stabilizing above 90\%.
Figure~\ref{fig:roc} shows ROC curves for all classes under a one-vs-rest setting. The proposed model achieves high AUC values on both datasets, with mean AUC of 0.965 for SMIDS and 0.991 for HuSHem, indicating excellent separability even under limited-data conditions.

\subsection{Explainable AI and Grad-CAM++ Visualizations}

\begin{figure}[h]
\centering
\setlength{\tabcolsep}{2pt}
\begin{tabular}{@{}c@{}c@{}}
\begin{tabular}{c}
SMIDS \\
\includegraphics[width=0.43\linewidth]{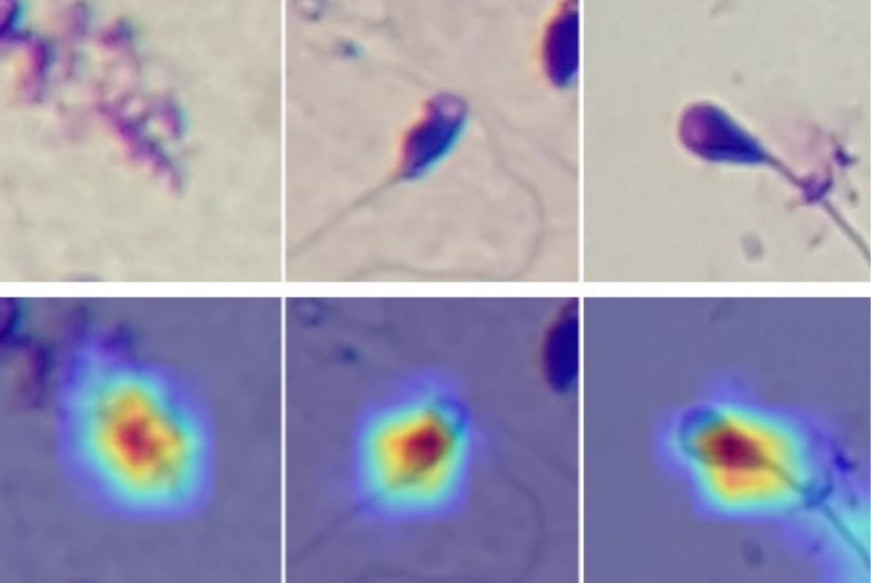}
\end{tabular}
&
\begin{tabular}{c}
HuSHem\\
\includegraphics[width=0.57\linewidth]{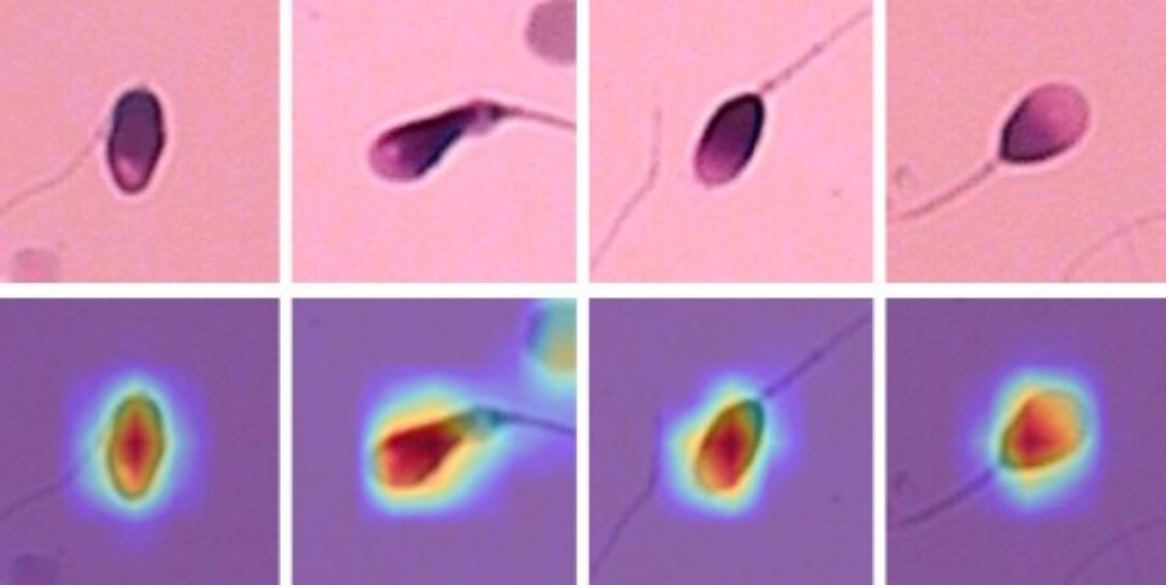}
\end{tabular}
\end{tabular}
\caption{Grad-CAM++ visualizations for SMIDS and HuSHem datasets. Each dataset is organized into two rows: original images (top) and explanation heatmaps (bottom).}
\label{fig:gradcam}
\end{figure}

To improve interpretability, Grad-CAM++ visualizations are generated from the final convolutional layer of the proposed network. Figure~\ref{fig:gradcam} presents representative results from both datasets.\\
The results show that the model consistently focuses on clinically relevant sperm head regions rather than background artifacts. In Normal samples, attention is concentrated on smooth oval head structures. In Abnormal, Pyriform, and Amorphous classes, attention shifts toward irregular boundaries and deformed regions. In Tapered sperm cells, the highest activation is localized at the elongated head tip, confirming that the model captures class-specific morphological patterns.\\
These observations confirm that CBAM improves spatial attention and enhances interpretability, making the decision process more transparent and clinically meaningful.

\subsection{Ablation Study}

The ablation study tracks how each component contributes to the overall success. Replacing SimpleCNN with a pretrained EfficientNet-B0 backbone improves SMIDS accuracy by 5.33\%, but reduces performance on HuSHem by 9.09\%. This suggests that directly fine-tuning a large pretrained model on only 150 training images leads to overfitting on background variations and staining artifacts.\\
Adding the CBAM attention module addresses this issue, improving HuSHem performance by 16.16\%, while slightly increasing SMIDS (+0.22\%). CBAM helps the model focus on the most relevant spatial locations and reweights feature channels toward the sperm head.\\
The freeze-then-unfreeze training strategy adds another 9.60\% gain on HuSHem. Starting with a frozen backbone allows the model to learn a stable initialization before full fine-tuning. Additionally, MixUp regularization contributes a 4.54\% improvement by smoothing decision boundaries, which is helpful in distinguishing between subtle morphology classes like Tapered and Pyriform. \\
Ultimately, the full model achieves a total improvement of 21.21\% on HuSHem and 7.54\% on SMIDS compared to the baseline. One limitation of this work is the small size of the HuSHem test set (33 samples), which introduces some uncertainty. Future studies should validate the method on larger, multi-centric datasets.

\begin{table}[h]
\centering
\scriptsize
\setlength{\tabcolsep}{2.5pt}
\renewcommand{\arraystretch}{1.1}
\caption{Ablation study: cumulative effect of adding each component to SimpleCNN baseline. $\Delta \nabla$ values show change relative to the previous row. Green = improvement, red = regression.}
\label{tab:ablation}
\begin{tabular}{lcccccc}
\toprule
\multirow{2}{*}{Configuration} 
& \multicolumn{2}{c}{SMIDS} 
&  & \multicolumn{2}{c}{HuSHem} 
& \\
\cmidrule(lr){2-3} \cmidrule(lr){5-6}
& Accuracy & F1-score & {\color{darkgreen}$\Delta$}{\color{darkred}$\nabla$} 
& Accuracy & F1-score & {\color{darkgreen}$\Delta$}{\color{darkred}$\nabla$} \\
\midrule
SimpleCNN 
& 82.67\% & 0.830 & --- 
& 72.73\% & 0.695 & --- \\
+ EfficientNet-B0 
& 88.00\% & 0.883 
& {\color{darkred}-5.33\%} 
& 63.64\% & 0.684 
& {\color{darkred}-9.09\%} \\
+ CBAM module 
& 88.22\% & 0.885 
& {\color{darkgreen}+0.22\%} 
& 79.80\% & 0.776 
& {\color{darkgreen}+16.16\%} \\
+ Freeze-unfreeze 
& 88.16\% & 0.883 
& {\color{darkred}-0.06\%} 
& 89.40\% & 0.882 
& {\color{darkgreen}+9.60\%} \\
Full model 
& 90.21\% & 0.913 
& {\color{darkgreen}+2.05\%} 
& 93.94\% & 0.948 
& {\color{darkgreen}+4.54\%} \\
\midrule
Total gain vs. Baseline 
& {\color{darkgreen}+7.54\%} & {\color{darkgreen}+0.083} & 
& {\color{darkgreen}+21.21\%} & {\color{darkgreen}+0.253} & \\
\bottomrule
\end{tabular}
\end{table}

\section{Conclusion}
Manual diagnosis of male infertility remains subjective and highly variable. This study addressed this limitation by proposing an automated framework that balances classification accuracy with clinical transparency. By combining a pretrained EfficientNet-B0 backbone with CBAM attention and a two-stage training strategy, our approach demonstrates high effectiveness, particularly when labeled medical data is scarce. The results on the small HuSHem dataset highlight a key lesson: while standard pretrained networks can overfit and perform worse than simple baselines on limited data, integrating spatial/channel attention and targeted fine-tuning completely reverses this regression. Furthermore, Grad-CAM++ visualizations confirm that the model's focus aligns with manual clinical criteria. Such interpretability is crucial for moving automated tools from research environments into real fertility laboratories, making transparency a core requirement rather than an afterthought.
\section*{Acknowledgements}
This work was supported by Vinnova, the Swedish Governmental Agency for Innovation Systems [Grant No. 2024-01462]. The funding source had no role in the design, execution, or publication of the study.

%-------------------------------------------------------------------
\clearpage
\bibliography{egbib}
\bibliographystyle{icml2026}

%-------------------------------------------------------------

\end{document}